\documentclass[10pt,twocolumn,letterpaper]{article}

\usepackage{iccv}
\usepackage{times}
\usepackage{epsfig}
\usepackage{graphicx}
\usepackage{amsmath}
\usepackage{amssymb}

\usepackage{subcaption}

\usepackage[pagebackref=true,breaklinks=true,letterpaper=true,colorlinks,bookmarks=false]{hyperref}

\iccvfinalcopy 

\def\iccvPaperID{1720} 

\ificcvfinal\fi
\begin{document}

\title{iQIYI-VID: A Large Dataset for Multi-modal Person Identification}

\author{Yuanliu~Liu,~Bo~Peng,~Peipei~Shi,~He~Yan,~Yong~Zhou,~Bing~Han,~Yi~Zheng,~Chao~Lin,\protect\\
        ~Jianbin~Jiang,Yin~Fan,~Tingwei~Gao,~Ganwen~Wang,~Jian~Liu,~Xiangju~Lu,~Danming~Xie\\
iQIYI, Inc.\\
}

\maketitle

\begin{abstract}
Person identification in the wild is very challenging due to great variation in poses, face quality, clothes, makeup and so on. Traditional research, such as face recognition, person re-identification, and speaker recognition, often focuses on a single modal of information, which is inadequate to handle all the situations in practice. Multi-modal person identification is a more promising way that we can jointly utilize face, head, body, audio features, and so on. In this paper, we introduce iQIYI-VID, the largest video dataset for multi-modal person identification. It is composed of 600K video clips of 5,000 celebrities. These video clips are extracted from 400K hours of online videos of various types, ranging from movies, variety shows, TV series, to news broadcasting. All video clips pass through a careful human annotation process, and the error rate of labels is lower than 0.2\%. We evaluated the state-of-art models of face recognition, person re-identification, and speaker recognition on the iQIYI-VID dataset. Experimental results show that these models are still far from being perfect for the task of person identification in the wild. We proposed a Multi-modal Attention module to fuse multi-modal features that can improve person identification considerably. We have released the dataset online to promote multi-modal person identification research.

\end{abstract}

\section{Introduction}



%
Nowadays, videos have dominated the flow on internet. Compared to still images, videos enrich the content by supplying audio and temporal information. As a result, video understanding is of urgent demand for practical usage, and person identification in videos is one of the most important tasks. Person identification has been widely studied in different research areas, including face recognition, person re-identification, and speaker recognition. In the context of video analysis, each research topic addresses a single modal of information. As deep learning arises in recent years, all these techniques have achieved great success. In the area of face recognition, ArcFace \cite{ArcFace} reached a precision of 99.83\% on the LFW benchmark \cite{LFWTech}, which had surpassed the human performance. The best results on Megaface \cite{MillerKS15} has also reached 99.39\%. For person re-identification (Re-ID), Wang \etal \cite{Wang_MM18} raised the Rank-1 accuracy of Re-ID to 97.1\% on the Market-1501 benchmark \cite{Market1501}. In the field of speaker recognition, the Classification Error Rates of SincNet \cite{SincNet} on the TIMIT dataset \cite{Garofolo93} and LibriSpeech dataset \cite{icassp/PanayotovCPK15} are merely 0.85\% and 0.96\%, respectively.

\begin{table*}
\caption{Datasets for person identification.}
\label{tab:dataset}
\centering
\begin{tabular}{lccccc} \hline
Dataset                                         & Task                  & Identities   & Format     & Clips/Face tracks & Images/Frames       \\ \hline
LFW \cite{LFWTech}                              & Face Recog.           & 5K           & Image      & -                 & 13K                 \\
Megaface \cite{MillerKS15}                      & Face Recog.           & 690K         & Image      & -                 & 1M                  \\
MS-Celeb-1M \cite{GuoZHHG16}                    & Face Recog.           & 100K         & Image      & -                 & 10M                 \\
YouTube Celebrities \cite{KimKPR08}             & Face Recog.           & 47           & Video      & 1,910             & -                   \\
YouTube Faces \cite{WolfHM11}                   & Face Recog.           & 1,595        & Video      & 3,425             & 620K                \\
Buffy the Vampire Slayer \cite{Sivic09}         & Face Recog.           & Around 19    & Video      & 12K               & 44K                 \\
Big Bang Theory \cite{BaumlTS13}                & Face\&Speaker Recog.  & Around 8     & Video      & 3,759             & -                   \\
Sherlock \cite{NagraniZ17}                      & Face\&Speaker Recog.  & Around 33    & Video      & 6,519             & -                   \\
VoxCeleb2 \cite{VoxCeleb2}                      & Speaker Recog.        & 6,112        & Video      & 150K              & -                   \\ \hline
Market1501 \cite{Market1501}                    & Re-ID                 & 1,501        & Image      & -                 & 32K                 \\
Cuhk03 \cite{CUHK03}                            & Re-ID                 & 1,467        & Image      & -                 & 13K                 \\
iLIDS \cite{iLIDS}                              & Re-ID                 & 300          & Video      & 600               & 43K                 \\
Mars \cite{MARS}                                & Re-ID                 & 1,261        & Video      & 20K               & -                   \\ \hline
CSM \cite{CSM}                                  & Search                & 1,218        & Video      & 127K              & -                   \\
iQIYI-VID                                       & Search                & 5,000        & Video      & 600K              & 70M                 \\
\hline
\end{tabular}
\end{table*}

Everything seems alright until we try to apply these person identification methods to the real unconstrained videos. Face recognition is sensitive to pose, blur, occlusion, \etc. Moreover, in many video frames the faces are invisible, which makes face recognition infeasible. When we turn to Re-ID, it has not touched the problem of changing clothes yet. For speaker recognition, one major challenge comes from the fact that the person to recognize is not always speaking. Generally speaking, every single technique is inadequate to cover all the cases. Intuitively, it will be beneficial to combine all these sub-tasks together, so we can fully utilize the rich content of videos.

There are several datasets for person identification in the literature. We list the popular datasets in Table \ref{tab:dataset}.
Most of the video datasets focus on only one modal of feature, either face \cite{KimKPR08,WolfHM11}, full body \cite{iLIDS,MARS,CSM}, or audio \cite{VoxCeleb2}. To our best knowledge there is no large-scale dataset that addresses the problem of multi-modal person identification.

Here we present the iQIYI-VID dataset, which is the first video dataset for multi-modal person identification. It contains 600K video clips of 5,000 celebrities, which is the largest celebrity identification dataset. These video clips are extracted from a huge amount of online videos in iQIYI\footnote{http://www.iqiyi.com/}. All the videos are manually labeled, which makes it a good benchmark to evaluate person identification algorithms. This dataset aims to encourage the development of multi-modal person identification.

To fully utilize different modal of features, we propose a Multi-modal Attention module (MMA) that learns to fuse multi-modal features adaptively according to their correlations. Compared to traditional ways of feature fusion such as Average Pooling or Concatenation, MMA can suppress the abnormal features that are inconsistent with other modal of features. In this way, we can handle the extreme cases that some modal of features are invalid, \eg the face is invisible or the character is not speaking.



\section{Related Work}

\subsection{Face Recognition}

In general, face recognition consists of two tasks, face verification and face identification. Face verification is to determine whether two given face images belong to the same person. In 2007, the Labeled Faces in the Wild (LFW) dataset was built for face verification by Huang \etal \cite{LFWTech}, which soon becomes the most popular benchmark for verification. Many algorithms \cite{FaceNet,DeepID3,SunWT15,DeepFace,TaigmanYRW15} have reached recognition rates of over 99\% on LFW, which is better than human performance \cite{HuYYKCLH15}. The state-of-art method, ArcFace \cite{ArcFace} achieved a face verification accuracy of 99.83\% on LFW. In this paper we take ArcFace to recognize faces in videos.

In recent years, the interest in face identification has greatly increased. The task of face identification is to find the most similar faces between gallery set and query set. Each individual in gallery set only has a few typical face images. Currently, the most challenging benchmarks are Megaface \cite{MillerKS15} and MS-Celeb-1M database \cite{GuoZHHG16}. Megaface includes one million unconstrained and multi-scaled photos of ordinary people collected from Flickr \cite{flickr}. The MS-Celeb-1M database provides one million face images of celebrities selected from free base with corresponding entity keys. Both datasets are large enough for training. However, their contents are still images. Moreover, these datasets are noisy as pointed out by \cite{DBLP:conf/eccv/WangCLHCQL18}. In 2008, Kim \etal created YouTube celebrity recognition database \cite{KimKPR08} that includes videos of only 35 celebrities, most of which are actors/actresses and politicians. The YouTube Face Database (YFD) \cite{WolfHM11} contains 3425 videos of 1595 different people. Both datasets are much smaller than iQIYI-VID. Another difference is that our dataset contains video clips without visible faces.

\subsection{Audio-based Speaker Identification}

Speaker recognition has been an active research topic for many decades \cite{Shaver16}. It comes in two forms that speaker verification \cite{ejasp/BimbotBFGMMMOPR04} aims to verify if two audio samples belong to the same speaker, whereas speaker identification \cite{Togneri2011AnOO} predicts the identity of the speaker given an audio sample. Speaker recognition has been approached by applying a variety of machine learning models \cite{SGBagul,ic3/BajpaiP09,SWFoo01}, either standard or specifically designed, to speech features such as MFCC \cite{conielecomp/MartinezPHS12}. For many years, the GMM-UBM framework \cite{dsp/ReynoldsQD00} dominated this particular field. Recently i-vector \cite{taslp/DehakKDDO11} and some related methodologies \cite{odyssey/Kenny10,icassp/CumaniPL13,Kenny05jointfactor} emerged and became increasingly popular. More recently, d-vector based on deep learning \cite{DeepSpeaker,icassp/HeigoldMBS16,icassp/VarianiLMMG14} had achieved competitive results against earlier approaches on some datasets. Nevertheless, speaker recognition still remains a challenge, especially for data collected in uncontrolled environment or from heterogeneous sources.

Currently, there are not many freely available datasets for speaker recognition, especially for large-scale ones. NIST has hosted several speaker recognition evaluations. However, the associated datasets are not freely available. There are datasets originally intended for speech recognition, such as TIMIT \cite{Garofolo93} and LibriSpeech \cite{icassp/PanayotovCPK15}, which have been used for speaker recognition experiments. Many of these datasets were collected under controlled conditions and therefore are improper for evaluating models in real-world conditions. To fill the gap, the Speaker in the Wild (SITW) dataset \cite{interspeech/McLarenFCL16} were created from open multi-media resources and freely available to the research community. To the best of our knowledge, the largest, freely available speaker recognition datasets are VoxCeleb \cite{VoxCeleb} and VoxCeleb2 \cite{VoxCeleb2}. To make sure that there are speakers in each clip, they added a key word `interview' to YouTube search. As a result the video scenes were limited to interviews. In comparison, iQIYI-VID is a multi-modal person identification dataset that covers more diverse scenes. Especially, there are video clips without speakers or with only asides.


\subsection{Body Feature Based Person Re-Identification}

Person re-identification (Re-ID) recognizes people across cameras, which is suitable for videos that have multiple shots switching between cameras. For single-frame-based methods, the model mainly extracted features from a still image, and directly determined whether the two pictures belong to the same person \cite{corr/GengWXT16,corr/LinZZWY17}. In order to improve the generalization ability, character attributes of the person were added to the network \cite{corr/LinZZWY17}. Metric learning measured the degree of similarity by calculating the distance between pictures, focusing on the design of the loss function \cite{eccv/VariorHW16,cvpr/SchroffKP15,corr/HermansBL17}. These methods relied on the global feature matching on the whole image, which were sensitive to the background. To solve this problem, local features gradually arose. Various strategies have been proposed to extract local features, including image dicing, skeleton point positioning, and attitude calibration \cite{eccv/VariorSLXW16,SpindleNet,AlignedReID}. Image dicing had the problem of data alignment \cite{eccv/VariorSLXW16}. Therefore, some researchers proposed to extract the pose information of the human body through skeleton point detection and introduce STN for correction \cite{SpindleNet}. Zhang \etal \cite{AlignedReID} further calculated the shortest path between local features and introduced a mutual learning approach, such that the accuracy of recognition exceeded that of humans for the first time.

Recently, video sequence-based methods are proposed to extract temporal information to aid Re-ID. AMOC \cite{corr/LiuJJQJYF17} adopted two sub-networks to extract the features of global content and motion. Song \etal \cite{aaai/SongLLHC18} adopted self-learning to find out low-quality image frames and reduce their importance, which made the features more representative.

\begin{figure*}[t]
  \centering
  \includegraphics[width=1.0\linewidth]{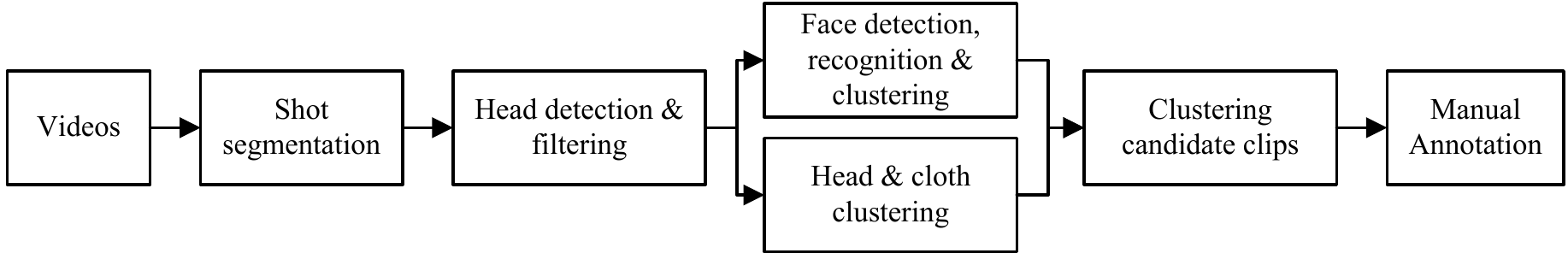}\\
  \caption{The process of building the iQIYI-VID dataset.}\label{fig:labeling}
\end{figure*}

Existing video-based Re-ID datasets, such as DukeMTMC \cite{eccv/RistaniSZCT16}, iLIDS \cite{iLIDS} and MARS \cite{MARS}, have invariant scenes and small amounts of data. The full bodies are usually visible, and the clothes are always unchanged for the same person across cameras. In comparison, video clips of iQIYI-VID are extracted from the massive video database of iQIYI, which covers various types of scenes and the same ID spans multiple kinds of videos. Significant changes in character appearances make Re-ID more challenging but closer to practical usage.

\section{iQIYI-VID DATASET}

The iQIYI-VID dataset is a large-scale benchmark for multi-modal person identification. To make the benchmark close to real applications, we extract video clips from real online videos of extensive types. Then we label all the clips by human annotators. Automatic algorithms are adopted to accelerate the collection and labeling process. The flow chart of the construction process is shown in Figure \ref{fig:labeling}. We begin with extracting video clips from a huge database of long videos. Then we filter out those clips with no person or multiple persons by automatic algorithms. After that we group the candidate clips by identity and put them into manual annotation. The details are given below.

\subsection{Extracting video clips}

The raw videos are the top 500,000 popular videos of iQIYI, covering movies, teleplays, variety shows, news broadcasting, and so on. Each raw video is segmented into shots according to the dissimilarity between consecutive frames. Video clips that are shorter than one second are excluded from the dataset, since they often lack multi-modal information. Clips longer than 30 seconds are also removed due to a large computation burden.

\subsection{Automatic filtering by head detection}
\label{sec:filter_by_head}
As a benchmark for person identification, each video clip is required to contain one and only one major character. In order to find out the major characters, heads are detected by YOLO V2 \cite{YOLO_V2}. A valid frame is defined as a frame in which only one head is detected, or the biggest head is three times larger than the other heads. A valid clip is defined as a clip whose valid frames exceed a ratio of 30\%. Invalid clips are removed. Since the head detector cannot detect all the heads, some clips will survive in this stage. Such noise clips will be thrown away in the manual filtering step.

\noindent\textbf{Discussion.} Taking clips with multiple persons inside will make the dataset closer to real applications, but it will largely increase the difficulty of annotation. Imaging a frame with a dozen of bit players in the background that can hardly be recognized by labelers.

\subsection{Obtaining candidate clips for each identity}

In this stage, each clip will be labeled with an initial identity by face recognition or clothes clustering. All the identities are selected from the celebrity database of iQIYI. The faces are detected by the SSH model \cite{SSH}, and then recognized by the ArcFace model \cite{ArcFace}. After that, the clothes and head information are used to cover those clips that no face has been detected or recognized. We cluster the clips from the same video by the faces and clothes information. The face and clothes in each frame are paired according to their relative position, and the identities of faces are propagated to the clothes with the face-clothes pairs. After that, each clothes cluster can get an ID through majority voting. The clips falling into the same cluster are regarded to be with the same identity, so the clips without any recognizable face can inherit the label from their clusters.

\subsection{Final manual filtering}

All clusters of clips are cleaned by a manual annotation process. We developed a video annotation system. In the annotation page, one part shows three reference videos that have high sores in face recognition, and the other part displays the video to be labeled. The labelers need to determine whether the video to be labeled has the same identity with the reference videos.


The manual labeling was repeated twice by different labelers to ensure a high quality. After data cleaning, we randomly selected 10\% of the dataset for quality testing and the data labeling error rate was kept within 0.2\%.
\section{Statistics}

\begin{figure}[t]
  \centering
    \begin{subfigure}[t]{0.45\columnwidth}
      \includegraphics[width=\textwidth]{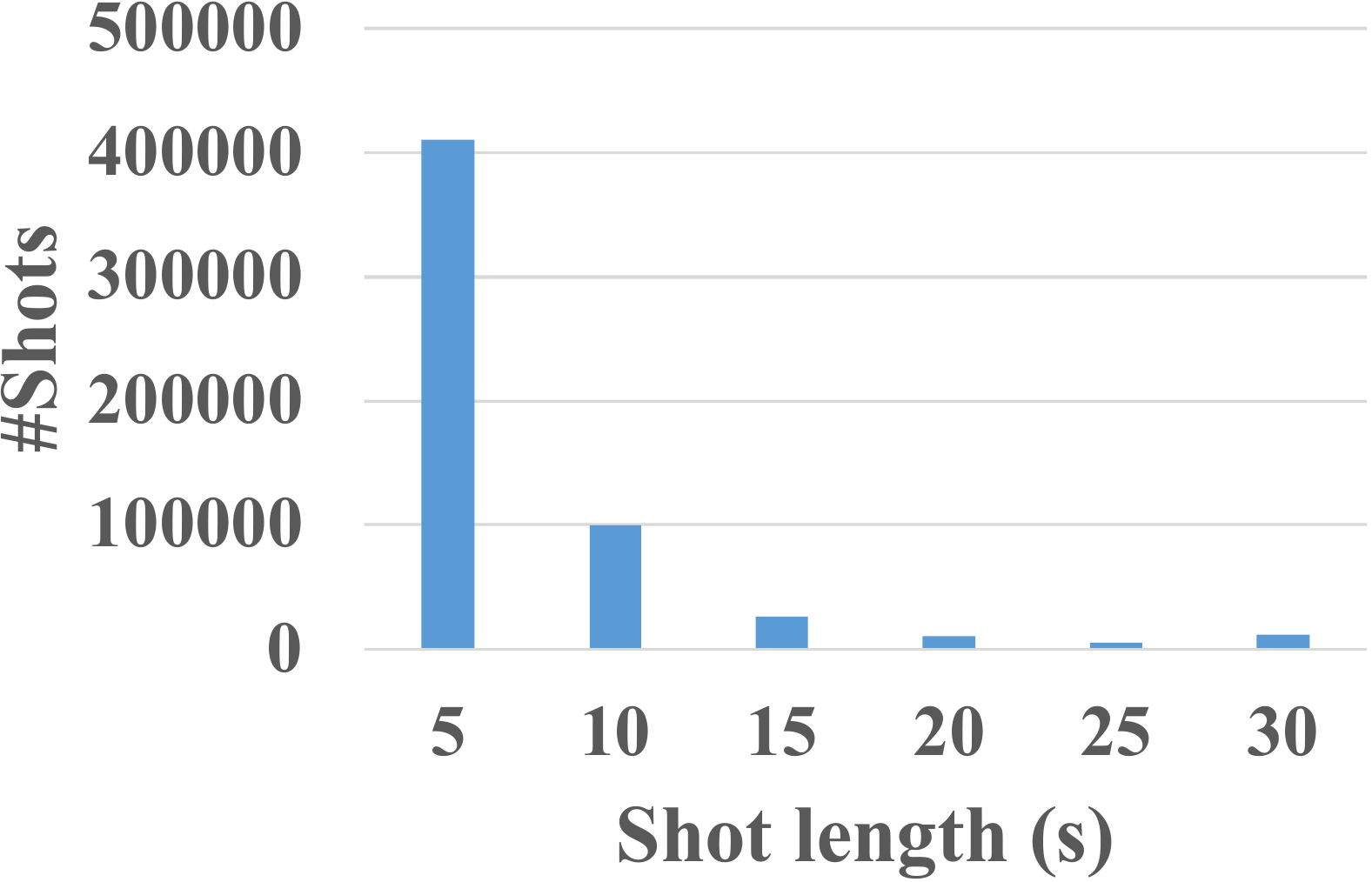}\\
      \caption{}
      \label{fig:Shot_length}
  \end{subfigure}
~
  \begin{subfigure}[t]{0.45\columnwidth}
      \includegraphics[width=\textwidth]{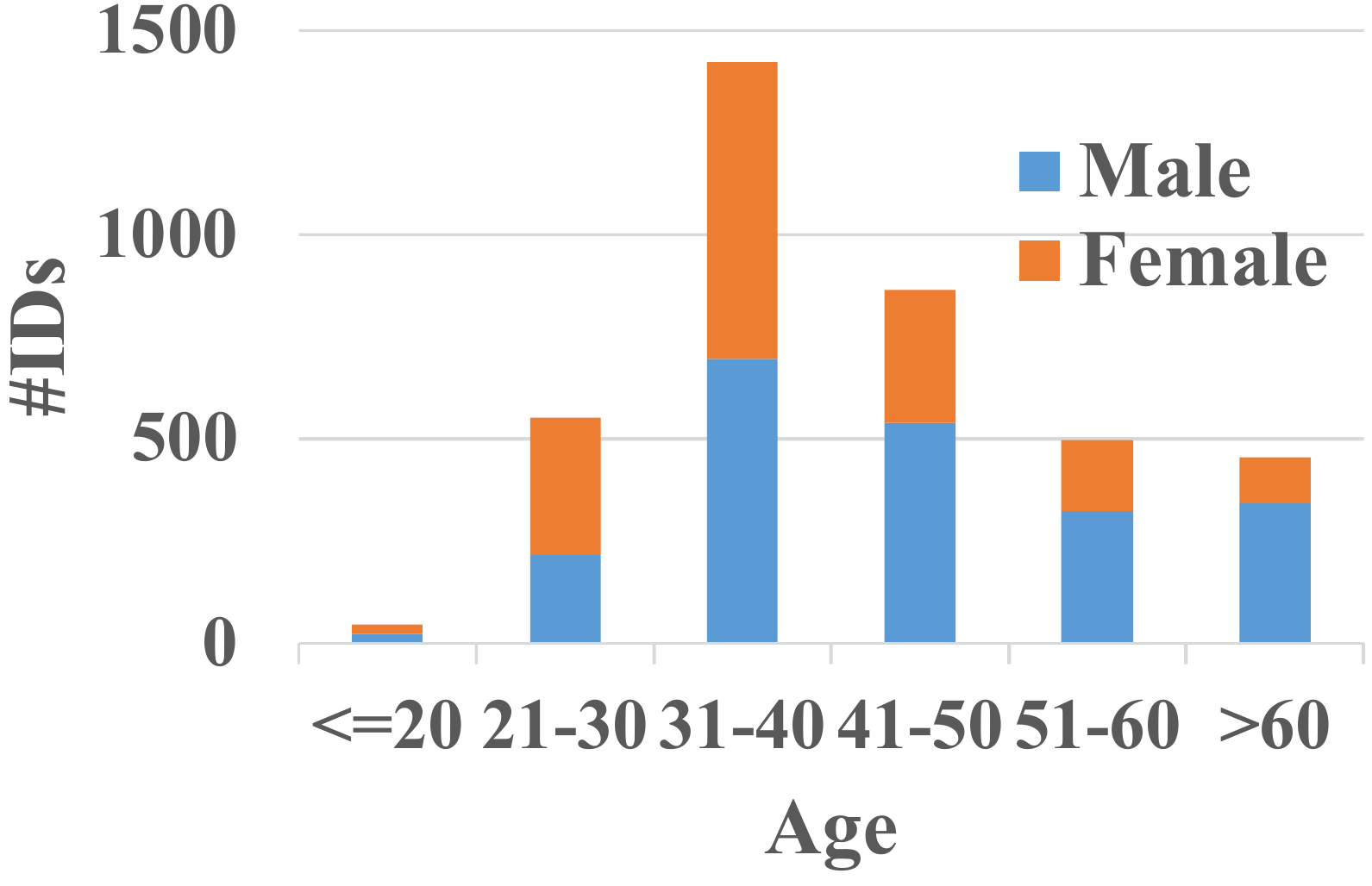}\\
      \caption{}
      \label{fig:age}
  \end{subfigure}
  ~
  \begin{subfigure}[b]{0.45\columnwidth}
      \includegraphics[width=\textwidth]{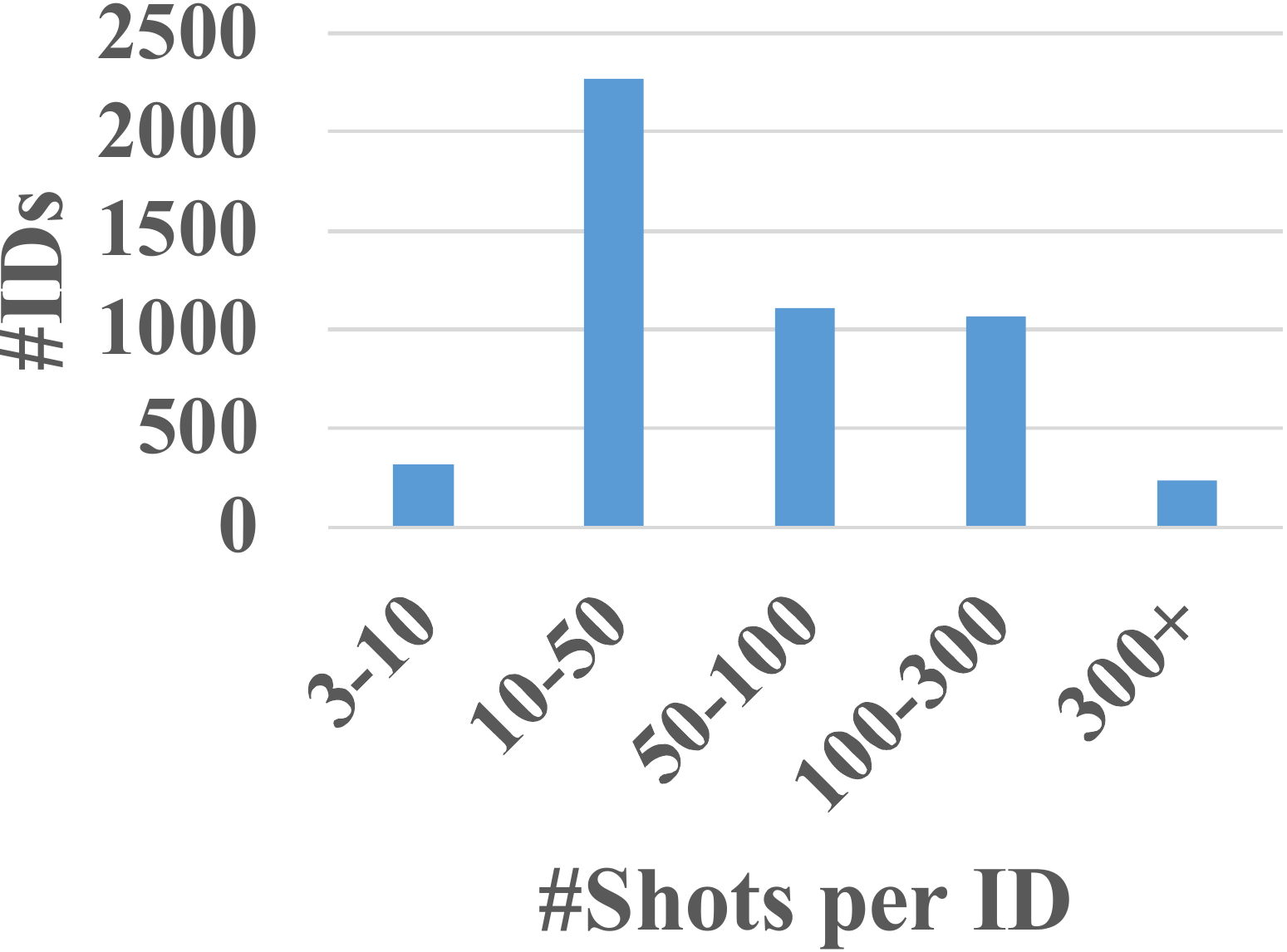}\\
      \caption{}
      \label{fig:Shots_per_ID}
  \end{subfigure}
  \caption{Data distributions.}\label{fig:video_age}
\end{figure}

\begin{figure*}[t]
  \centering
  \includegraphics[width=0.9\linewidth]{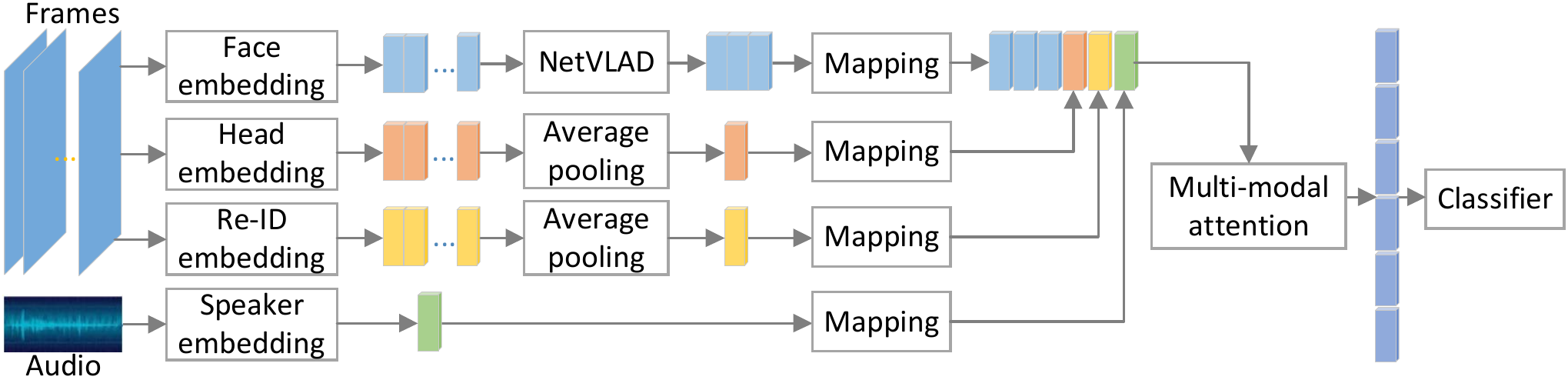}\\
  \caption{The flowchart of our multi-modal person identification method. It begins with extracting raw features of face, head, body, and audio. The raw features of face, head, and body are transformed into video-level features by an adapted NetVLAD module or Average Pooling. Then they are transformed to the same length by feature mapping. Different modal of features are combined by multi-modal attention, and then fed to a classifier.}\label{fig:model}
\end{figure*}


The iQIYI-VID dataset contains 600K video clips. The whole dataset is divided into three parts, 40\% for training, 30\% for validation, and 30\% for test. Researchers can download the training and validation parts from the website \footnote{http://challenge.ai.iqiyi.com/detail?raceId=5afc36639689443e8f815f9e}, and the test part is kept secret for evaluating the performance of contesters. The video clips are labeled with 5,000 identities from the iQIYI celebrity database. Among all the celebrities, 4,360 of them are Asian, 510 are Caucasian, 41 are African, and 23 are Hispanic. The primary language of speech is Chinese. To mimic the real environment of video understanding, we add 84,759 distracter videos with unknown person identities (outside the 5,000 major identities in training set) into the validation set and another 84,746 distracter videos into the test set.

The number of video clips for each celebrity varies between 20 and 900, with 100 on average. The histogram of ID over the number of video clips is shown in Figure \ref{fig:Shots_per_ID}. As shown in Figure \ref{fig:Shot_length}, the video clip duration is in the range of 1 to 30 seconds, with 4.72 seconds on average. There are 791.6 hours of video clips in total.

The dataset is approximately gender-balanced, with 54\% males and 46\% females. The age range of the dataset is large. The earliest birth date was 1877 and the latest birth date was 2012. The histogram is shown in Figure \ref{fig:age}.



\section{Baseline Approaches}
\label{sec:models}
In this section we present a baseline approach for multi-modal person identification on the iQIYI-VID dataset. The flowchart is shown in Figure \ref{fig:model}. Multi-modal raw features are extracted from single frames or audio. Then the frame-level features are transformed into video-level features by a NetVLAD module or Average Pooling. The video-level features are mapped into new feature spaces that have the same dimension, so they can be concatenated into a feature map along the dimension of modality. A multi-modal attention module is proposed to fuse different modal of features together. The fused feature is fed to a classifier that predicts the ID of the video clip. For evaluation, video retrieval is realized by ranking test videos according to the probability distribution over the IDs.


\subsection{Raw features}
\label{raw_features}

Our method uses four modal of features, including face, head, body, and audio. Raw features are extracted by the state-of-art models, as described below.

\noindent\textbf{Face}. We use the SSH model \cite{SSH} to detect faces, which is one of the best open-source face detectors. To accelerate the detection, we replace the VGG16 backbone network \cite{VGG} of the original SSH with MobileNetV1 \cite{MobileNetV1}. For each face, we utilize the state-of-art face recognition model ArcFace \cite{ArcFace} for feature extraction.

Face is a critical clue for person identification when and only when the face is visible and the face quality is acceptable to the face classifier. To pick up those high-quality faces from the video clip, we simply rank the faces by the L2-norm of the output of the FC1 layer in the face recognition model of SphereFace as our face quality score \cite{cvpr/LiuWYLRS17}. Ranjan et al. observed that the L2-norm of the features learned using softmax loss is informative for measuring face quality \cite{RanjanCC17}. In our experiments, the faces that are regarded as low-quality are mostly those blurred faces, side faces, and partially misdetected faces. We keep the top 32 faces for feature fusion in the next stage. When there are less than 32 faces in the video clip, we randomly sample the existing faces until the number of faces reaches 32.

\noindent\textbf{Head}. The heads are detected by YOLO V2 \cite{YOLO_V2} as mentioned in Section \ref{sec:filter_by_head}. We train a head classifier based on the ArcFace model \cite{ArcFace}. The head features contain information from hair style and accessories, which are good supplements to faces.

\noindent\textbf{Body}. We detect persons by a SSH detector \cite{SSH} trained on the CrowdHuman dataset \cite{crowdhuman}. We utilize Alignedreid++ \cite{AlignedReID++} to recognize persons by their body features.

\noindent\textbf{Audio}. The audio from the video clips is converted to a single-channel, 16-bit signal with a 16kHz sampling rate. The magnitude spectrograms of these audio are mean-subtracted, but variance normalisation is not performed. Instead, we only divide those mean-subtracted spectrograms by a constant value 50.0 and feed the results as input to a CNN model based on ResNet34 \cite{he2016deep}, inspired by  \cite{VoxCeleb}. The CNN model is trained as a classification model using the Dev part of Voxceleb2 dataset \cite{ic3/BajpaiP09} with 5994 speakers, $14\%$ of the data is used as evaluation while the rest as training data. We achieved a best classification error rate of $6.3\%$ on the evaluation data. The 512D output from the last hidden layer is used as speaker embedding.

\subsection{Video-level features}
\label{video_level_features}

\begin{figure}[tb]
  \centering
  \includegraphics[width=1.0\linewidth]{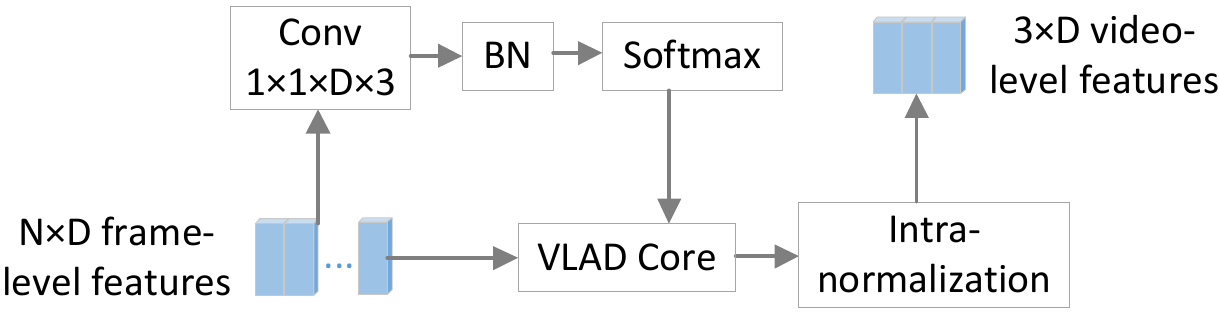}\\
  \caption{Adapted NetVLAD module. A Batch Normalization (BN) layer is added to the Fully-connected (FC) layer. The \textbf{VLAD core} layer is composed of a subtraction of the cluster center and a weighted sum. Refer to \cite{NetVLAD} for more details.}\label{fig:NetVLAD}
\end{figure}

The raw features extracted in Section \ref{raw_features} cannot be used directly for person recognition, since there may be too many frames in a video clip and the order and amount of frames are quite different from clip to clip. We have to aggregate the frame-level features into video-level features of a regularized form, while keeping the network trainable. NetVLAD \cite{NetVLAD} is a widely used module for extracting mid-level features. In our experiments we found that the convergence of the model is quite slow during training when using the original NetVLAD, so we add a Batch-normalization (BN) layer after the Fully-connected (FC) layer to accelerate training. We also find it is better to remove the L2 norm in our case. The structure of the adapted NetVLAD module is shown in Figure \ref{fig:NetVLAD}. We use a FC layer to map all the video-level features into the same dimension.


\subsection{Feature fusion}
\label{sec:fusion}

\begin{figure}[tb]
  \centering
  \includegraphics[width=1.0\linewidth]{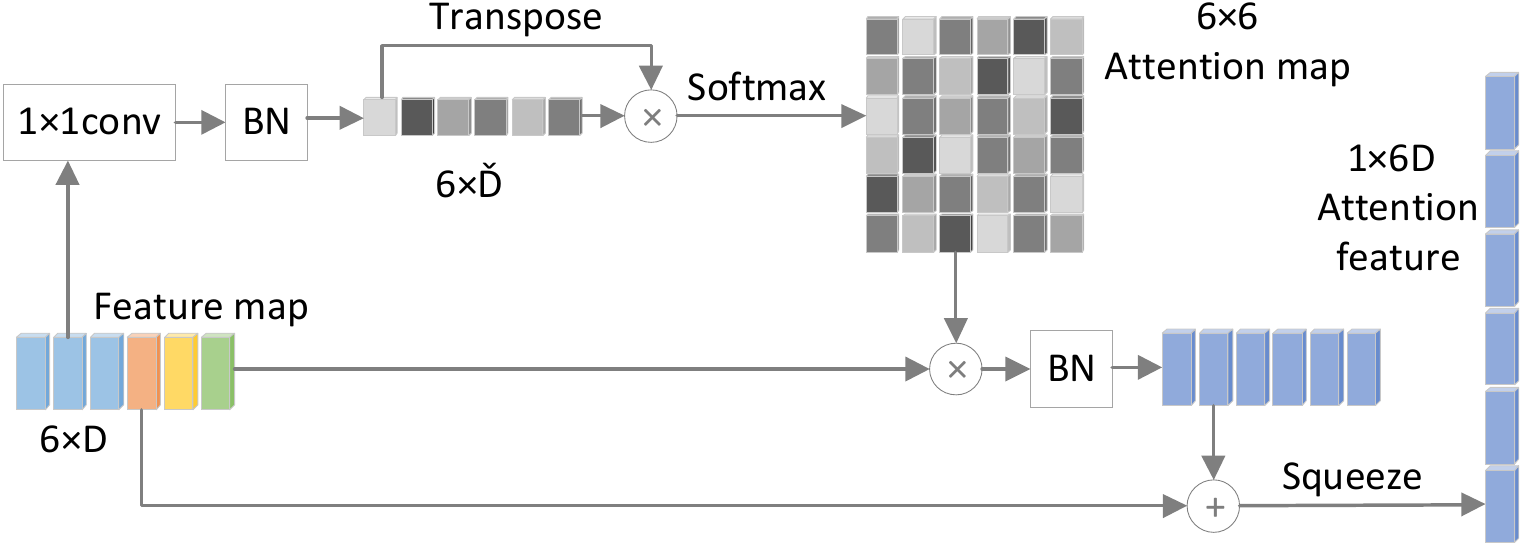}\\
  \caption{Multi-modal attention (MMA) module.}\label{fig:MMA}
\end{figure}

We propose a Multi-modal Attention module (MMA) to fuse different modal of features together, as shown in Figure \ref{fig:MMA}. The input of MMA is a feature map $X\in\mathbb{R}^{D\times 6}$, where 3 of the 6 features come from face, and the other 3 features come from head, body, and audio, as shown in Figure \ref{fig:model}. Each feature has the length of $D$. The feature map $X$ is transformed into a feature space $F\in\mathbb{R}^{\check{D}\times 6}$ by $W_F\in\mathbb{R}^{\check{D}\times D}$ to calculate the attention $Y\in\mathbb{R}^{6\times 6}$, where
\begin{equation}
    Y_{i,j} = \frac{\exp(Z_{i,j})}{\sum_{i=1}^{6}\exp(Z_{i,j})},
\end{equation}
with
\begin{equation}
    Z=F^{\top}F, \ F=W_FX.
\end{equation}
Here $Z$ is the Gram matrix of feature map $F$, which captures the feature correlation \cite{nips/GatysEB15} that is widely used for image style transfer \cite{cvpr/GatysEB16}.

The fused feature map $O\in\mathbb{R}^{D\times 6}$ is obtained by
\begin{equation}
    O=X+\gamma XY,
\end{equation}
where $\gamma$ is a scalar parameter to control the strength of attention. At last the fused feature map is squeezed into a vector and fed to the classifier.

%

\noindent\textbf{Discussion}. MMA is inspired by the Self-Attention module used in SAGAN \cite{SelfAttention}. The major difference is that Self-Attention module captures long-range \textbf{spatial dependencies} in images, while MMA aims at learning \textbf{inter-modal correlation}. After fusion by MMA, a modal of feature that is consistent with other modal of feature will be amplified while the inconsistent features will be suppressed. In another point of view, MMA will "smooth" different modal of features, which is important to compensate the loss when some features are missing, \eg, the video has no audio channel. Another difference is that Self-Attention module maps the input feature map into two individual feature spaces and measure the correlation by their transposed multiplication. Instead we adopt the Gram matrix of a unique feature map since it is simpler and more effective in capturing the feature correlation. In practical we also found that the Gram matrix performs better.


\section{Experiments}

\subsection{Experiment setup}

 The modified SSH face detector is trained on the Widerface dataset \cite{yang2016wider}. In Section \ref{sec:fusion}, the feature length $D$ is set to 512, and the transformed feature length $\check{D}$ is set to 64. The weight $\gamma$ is set to 0.05.
\subsection{Evaluation Metrics}


To evaluate the retrieval results, we use Mean Average Precision ($MAP$) \cite{IR}:
\begin{equation}
\label{eqn:map}
MAP(Q) = \frac{1}{|Q|}\sum_{i=1}^{|Q|}\frac{1}{m_i}\sum_{j=1}^{n_i}Precision(R_{i,j}),
\end{equation}
where $Q$ is the set of person IDs to retrieve, $m_i$ is the number of positive examples for the $i$-th ID, $n_i$ is the number of positive examples within the top $k$ retrieval results for the $i$-th ID, and $R_{i,j}$ is the set of ranked retrieval results from the top until you get $j$ positive examples. In our implementation, only top 100 retrievals are kept for each person ID. Note that, we did not use the top K accuracy in evaluation since we added many video clips of unknown identities to the test set, which makes the top K accuracy invalid.

\begin{table}
\caption{Comparison to the state-of-art.}
\label{tab:state_of_art}
\centering
\begin{tabular}{lc} \hline
Modal                               & MAP (\%)    \\ \hline
ArcFace \cite{ArcFace}              & 88.65       \\
He \etal \cite{iQIYI_challenge_res} & 89.00       \\
Ours                                & 89.70       \\
\hline
\end{tabular}
\end{table}

\begin{table}
\caption{Results of different modal of features and modules.}
\label{tab:results_multi_modal}
\centering
\begin{tabular}{lcccc} \hline
Modal                               & MAP (\%)    \\ \hline
Face                                & 85.19       \\
Head                                & 54.32       \\
Audio                               & 11.79       \\
Body                                & 5.14        \\ \hline
Face+Head                           & 87.16       \\
Face+Head+Audio                     & 87.69       \\
Face+Head+Audio+Body                & 87.80       \\ \hline
Ensemble                            & 89.24       \\ \hline
+NetVLAD                            & 89.46       \\
+NetVLAD+MMA                        & 89.70       \\
\hline
\end{tabular}
\end{table}

\subsection{Results on iQIYI-VID}

\begin{figure}[htb]
  \centering
  \includegraphics[width=0.9\linewidth]{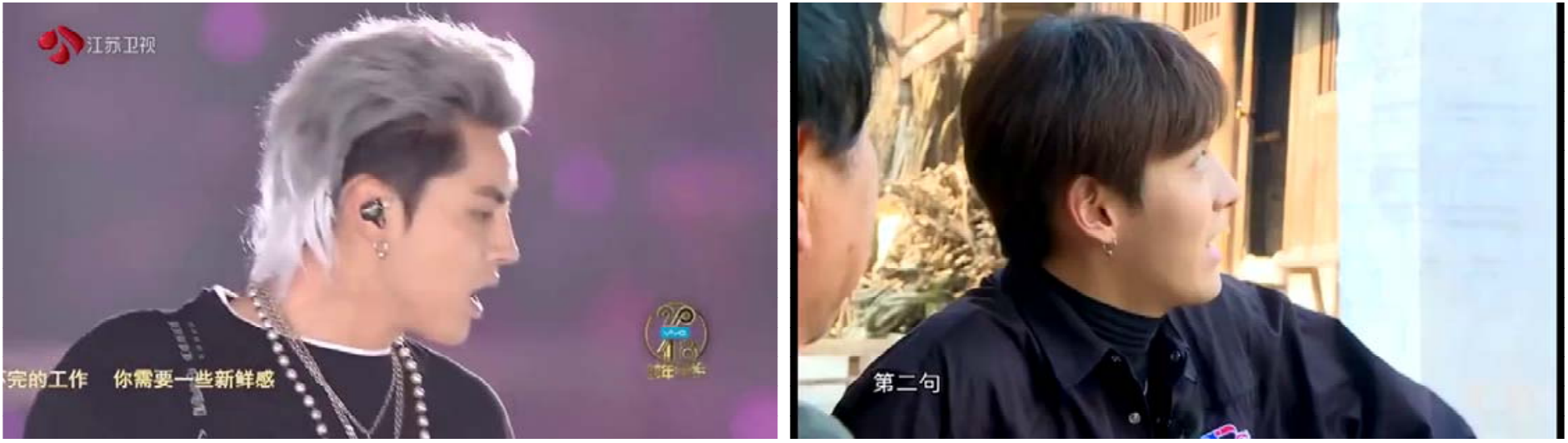}\\
  \caption{Challenging cases for head recognition. The hair style and accessories of the same actor changed dramatically.}\label{fig:head_hard}
\end{figure}

\begin{figure}[htb]
  \centering
  \includegraphics[width=0.9\linewidth]{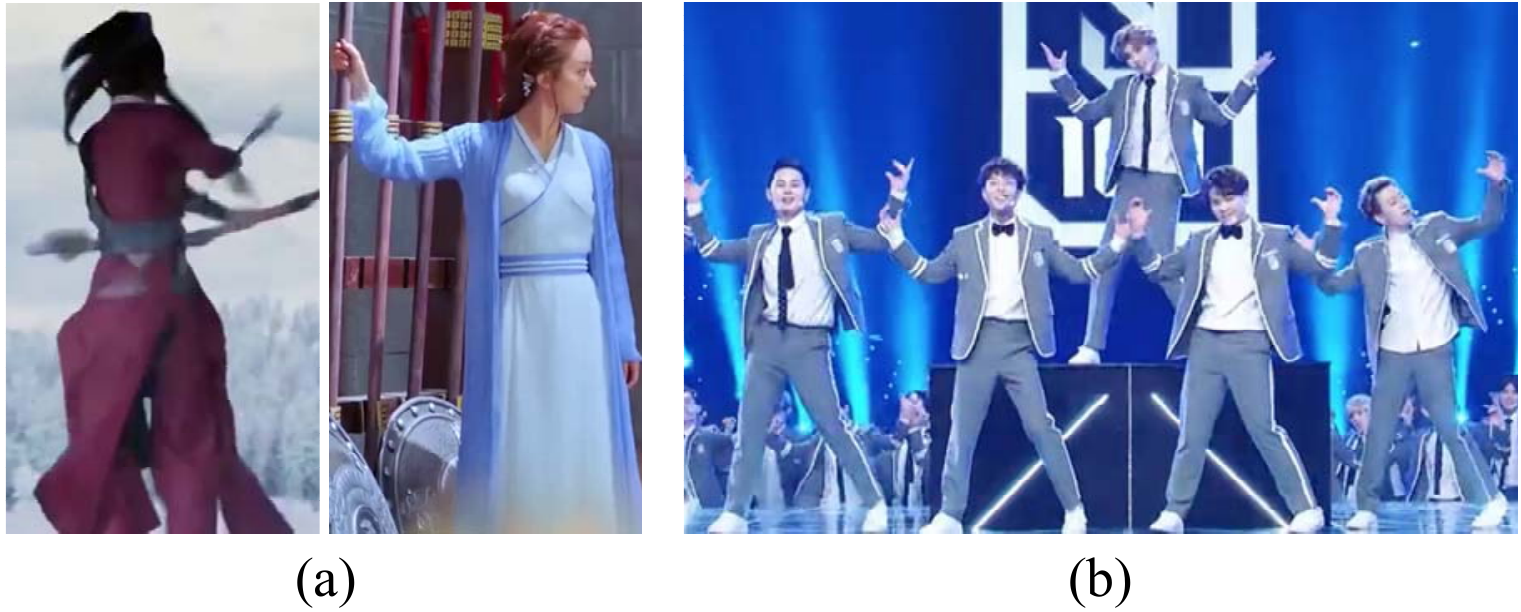}\\
  \caption{Challenging cases for body recognition. (a) An actress changes style in different episodes. (b) Different actors dress the same uniform. }\label{fig:body_hard}
\end{figure}

\begin{figure*}[t]
  \centering
  \includegraphics[width=0.9\linewidth]{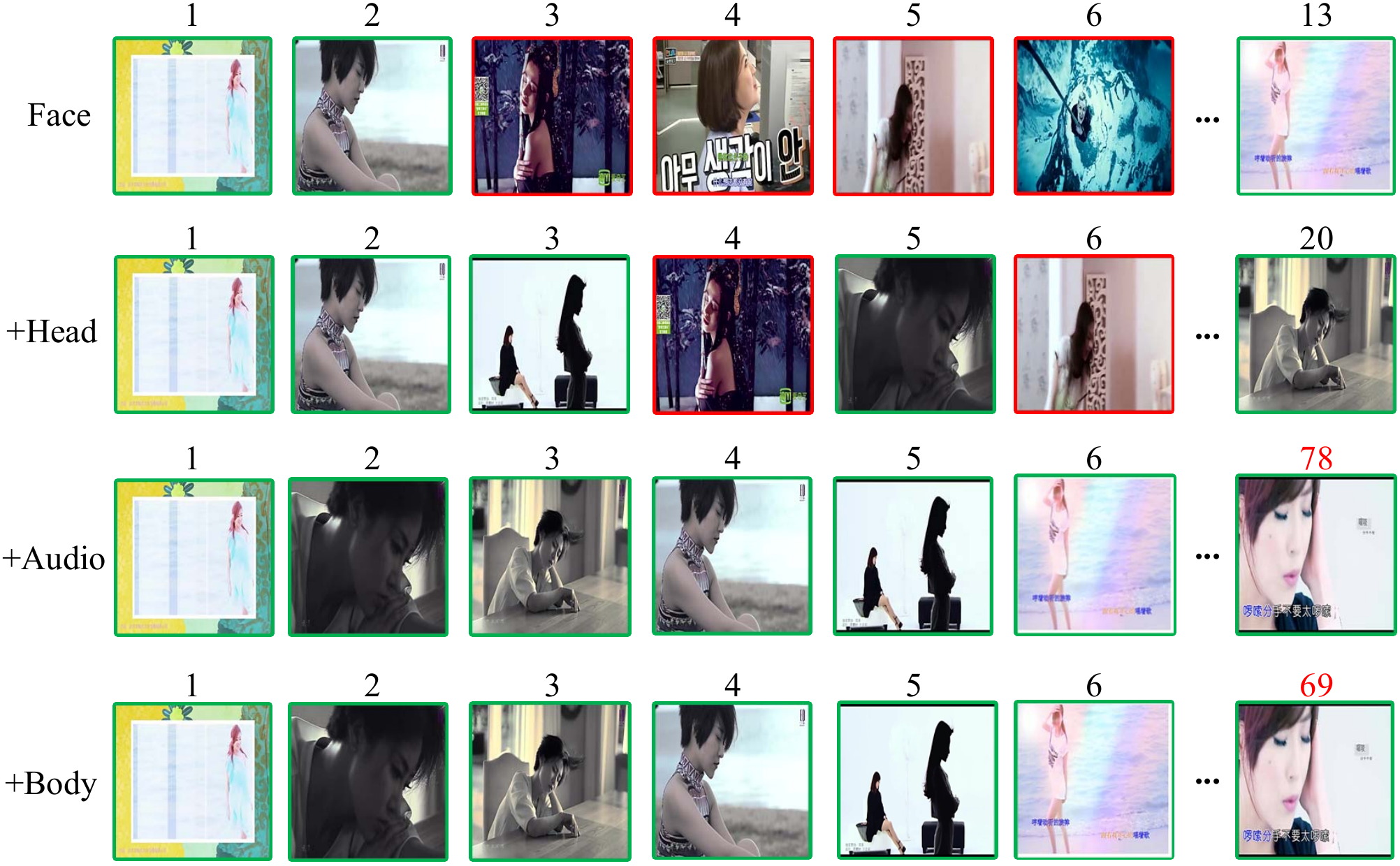}\\
  \caption{An example of video retrieval results. When adding more and more features inside, the results get much better than face recognition alone. The number above each image indicates the rank of the image within the retrieval results. Positive examples are marked by green boxes, while negative examples are marked in red.}\label{fig:results_multi_modal}
\end{figure*}

\begin{figure*}[htb]
  \centering
  \includegraphics[width=0.9\linewidth]{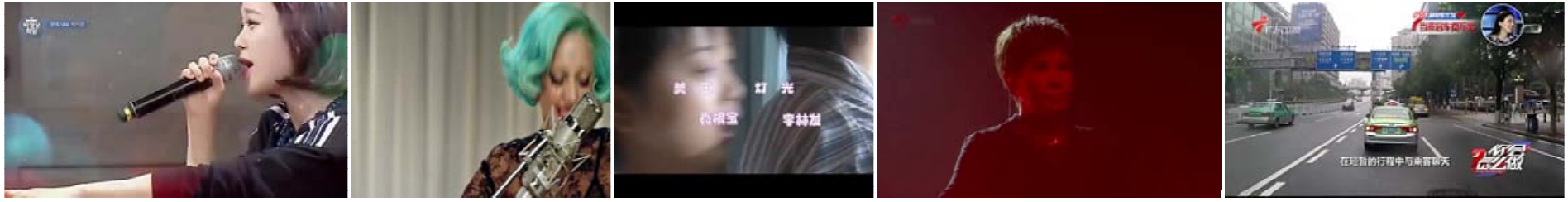}\\
  \caption{Challenging cases for face recognition. From left to right: profile, occlusion, blur, unusual illumination, and small face.}\label{fig:face_rec_hard}
\end{figure*}

We evaluate the models described in Section \ref{sec:models} on the test set of iQIYI-VID dataset. We compare to the state-of-art methods in Table \ref{tab:state_of_art}. Our method achieved a MAP of 89.70\%, which was 0.70\% higher than the current state-of-art. ArcFace \cite{ArcFace} adopted raw features trained on MS1MV2 and Asian datasets, and fed them to a Multi-layer Perceptron. Besides, they combined six models together in a cross-validation fashion, and took the context features from off-the-shelf object and scene classifiers. He \etal \cite{iQIYI_challenge_res} randomly merged several classes into one meta-class and train 7 meta-class classifiers with triplet-loss. The final prediction was given by calculating the cosine distance of the concatenated features from the 7 meta-class classifiers.

\subsection{Ablation study}

In this section we evaluate the effects of multi-modal features as well as the NetVLAD and Multi-modal Attention (MMA) modules on our model. The results are given in Table \ref{tab:results_multi_modal}. The basic method takes the face features as input, combining the frame-level feature by a simple Average Pooling, and removing the MMA module.

\noindent\textbf{Multi-modal attention}. Among all the models using single modal of features, face recognition achieved the best performance. It should be mentioned that, ArcFace \cite{ArcFace} achieved a precision of $99.83\%$ on the LFW benchmark \cite{LFWTech}, which is much higher than on the iQIYI-VID dataset. It suggested that the iQIYI-VID dataset is more challenging than LFW. In Figure \ref{fig:face_rec_hard} we show some difficult examples that are hard to recognize using only face features.

None of the other features alone is comparable to face features. The head feature is unreliable when the face is hidden. In this case the classifier mainly relies on hair or accessories, which are much less discriminative than face features. Moreover, the actors or actresses often change their hairstyles across the shows, as shown in Figure \ref{fig:head_hard}.

The MAP of audio-based model is only $11.79\%$. The main reason is that the person identity of the video clip is primarily determined by the figure in the video frame, and we did not use a voice detection module to filter out non-speaking clips, nor employ an active speaker detection. As a result, the sound may likely not come from the person of interest. Moreover, in many cases, even though the person of interest does speak, the voice actually comes from a voice actor for that person, which makes recognition by speaker more problematic. When the character does not say anything inside the clip and the audio may come from the background, or even from some other characters, which are distracters for speaker recognition.

The performance of body feature is even worse. The main challenge comes from two aspects. In the one hand, the clothes of the characters always change from one show to another. In the other hand, the uniforms of different characters may be nearly the same in some shows. Figure \ref{fig:body_hard} gives an example. As a result, the intra-class variation is comparable to, if not larger than, the inter-class variation.

Although neither head, audio, nor body feature alone can recognize person well enough, they can produce better classifiers when combined with the face feature. From Table \ref{tab:results_multi_modal} we can see that, adding more features always achieves better performances. Adopting all four kinds of features can raise the MAP by over $2.61\%$, from $85.19\%$ to $87.80\%$. It proves that multi-modal information is important for video-based person identification. An example is shown in Figure \ref{fig:results_multi_modal}. We can see that when multi-modal features are added to person recognition gradually, more and more positive examples are retrieved back.

\noindent\textbf{Ensemble}. Inspired by the strategy of Cross Validation, we built a simple expert system that is an ensemble of models trained on different partition of the training set. The output of the expert system is the sum of the output probability distribution of all the models. When using three models, the MAP can be raised by $1.44\%$, which is a considerable margin. 

\noindent\textbf{NetVLAD}. Replacing Average Pooling by NetVLAD for the face feature can increase the MAP by $0.22\%$, which proves the efficacy of a good mid-level feature. Note that, we did not apply NetVLAD to the other modal of features, since they have minor impact on the final results while the computation burden would be largely raised.

\begin{figure}[htb]
  \centering
  \includegraphics[width=0.4\linewidth]{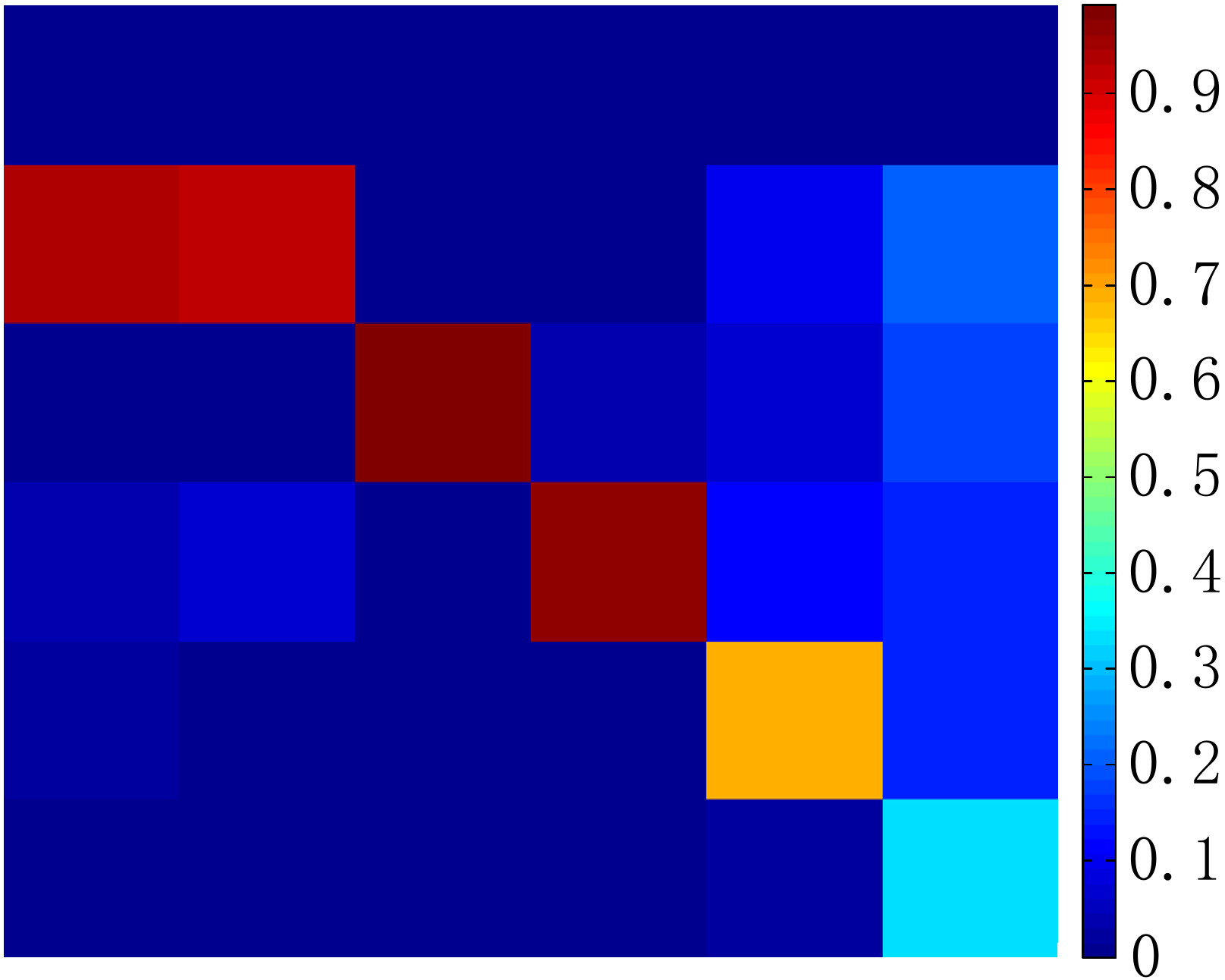}\\
  \caption{The average Gram matrix of MMA module on test set. Column $i$ represents the correlation of Feature $i$ to each of the 6 features. }\label{fig:Gram}
\end{figure}

\noindent\textbf{Multi-modal attention}. MMA module can increase the MAP by $0.24\%$, which is a noticeable margin. Figure \ref{fig:Gram} shows the average Gram matrix of MMA module on test set. It suggests that the first face feature is largely replaced by the second face feature, since they are highly correlated and the second face feature more favorable. The audio feature distributed most of its weights to the other features, which is consistent with the fact that audio feature is unstable.

\section{CONCLUSIONS AND FUTURE WORK}

In this work, we investigated the challenges of person identification in real videos.  We built a large-scale video dataset called iQIYI-VID, which contains more than 600K video clips and 5,000 celebrities extracted from iQIYI copyrighted videos. We proposed a MMA module to fuse different modal of features adaptively according to their correlation. Our baseline approach of multi-modal person identification demonstrated that it is beneficial to take different sources of features to deal with real-world videos. We hope this new benchmark can promote the research in multi-modal person identification.


{\small

\begin{thebibliography}{10}\itemsep=-1pt

\bibitem{flickr}
Flickr.
\newblock \url{https://www.flickr.com/}.

\bibitem{iQIYI_challenge_res}
Latest rank of iqiyi celebrity video identification challenge.
\newblock \url{http://challenge.ai.iqiyi.com/}.

\bibitem{NetVLAD}
R.~Arandjelovic, P.~Gron{\'{a}}t, A.~Torii, T.~Pajdla, and J.~Sivic.
\newblock Netvlad: {CNN} architecture for weakly supervised place recognition.
\newblock In {\em CVPR}, pages 5297--5307, 2016.

\bibitem{SGBagul}
S.~G. Bagul and R.~K. Shastri.
\newblock Text independent speaker recognition system using gmm.
\newblock In {\em International Conference on Human Computer Interactions},
  pages 1--5, 2013.

\bibitem{ic3/BajpaiP09}
A.~Bajpai and V.~Pathangay.
\newblock Text and language-independent speaker recognition using
  suprasegmental features and support vector machines.
\newblock In {\em International Conference on Contemporary Computing}, pages
  307--317, 2009.

\bibitem{BaumlTS13}
M.~B{\"{a}}uml, M.~Tapaswi, and R.~Stiefelhagen.
\newblock Semi-supervised learning with constraints for person identification
  in multimedia data.
\newblock In {\em CVPR}, pages 3602--3609, 2013.

\bibitem{ejasp/BimbotBFGMMMOPR04}
F.~Bimbot, J.~Bonastre, C.~Fredouille, G.~Gravier, I.~Magrin{-}Chagnolleau,
  S.~Meignier, T.~Merlin, J.~Ortega{-}Garcia, D.~Petrovska{-}Delacr{\'{e}}taz,
  and D.~A. Reynolds.
\newblock A tutorial on text-independent speaker verification.
\newblock {\em {EURASIP} J. Adv. Sig. Proc.}, 2004(4):430--451, 2004.

\bibitem{VoxCeleb2}
J.~S. Chung, A.~Nagrani, and A.~Zisserman.
\newblock Voxceleb2: Deep speaker recognition.
\newblock In {\em Interspeech}, pages 1086--1090, 2018.

\bibitem{icassp/CumaniPL13}
S.~Cumani, O.~Plchot, and P.~Laface.
\newblock Probabilistic linear discriminant analysis of i-vector posterior
  distributions.
\newblock In {\em International Conference on Acoustics, Speech and Signal
  Processing}, pages 7644--7648, 2013.

\bibitem{taslp/DehakKDDO11}
N.~Dehak, P.~Kenny, R.~Dehak, P.~Dumouchel, and P.~Ouellet.
\newblock Front-end factor analysis for speaker verification.
\newblock {\em {IEEE} Trans. Audio, Speech {\&} Language Processing},
  19(4):788--798, 2011.

\bibitem{ArcFace}
J.~Deng, J.~Guo, N.~Xue, and S.~Zafeiriou.
\newblock Arcface: Additive angular margin loss for deep face recognition.
\newblock In {\em CVPR}, 2019.

\bibitem{SWFoo01}
S.~W. Foo and E.~G. Lim.
\newblock Speaker recognition using adaptively boosted classifier.
\newblock In {\em International Conference on Electrical and Electronic
  Technology}, 2001.

\bibitem{nips/GatysEB15}
L.~A. Gatys, A.~S. Ecker, and M.~Bethge.
\newblock Texture synthesis using convolutional neural networks.
\newblock In {\em NIPS}, pages 262--270, 2015.

\bibitem{cvpr/GatysEB16}
L.~A. Gatys, A.~S. Ecker, and M.~Bethge.
\newblock Image style transfer using convolutional neural networks.
\newblock In {\em CVPR}, pages 2414--2423, 2016.

\bibitem{corr/GengWXT16}
M.~Geng, Y.~Wang, T.~Xiang, and Y.~Tian.
\newblock Deep transfer learning for person re-identification.
\newblock {\em CoRR}, abs/1611.05244, 2016.

\bibitem{GuoZHHG16}
Y.~Guo, L.~Zhang, Y.~Hu, X.~He, and J.~Gao.
\newblock Ms-celeb-1m: {A} dataset and benchmark for large-scale face
  recognition.
\newblock In {\em European Conference on Computer Vision}, pages 87--102, 2016.

\bibitem{he2016deep}
K.~He, X.~Zhang, S.~Ren, and J.~Sun.
\newblock Deep residual learning for image recognition.
\newblock In {\em Proceedings of the IEEE conference on computer vision and
  pattern recognition}, pages 770--778, 2016.

\bibitem{icassp/HeigoldMBS16}
G.~Heigold, I.~Moreno, S.~Bengio, and N.~Shazeer.
\newblock End-to-end text-dependent speaker verification.
\newblock In {\em International Conference on Acoustics, Speech and Signal
  Processing}, pages 5115--5119, 2016.

\bibitem{corr/HermansBL17}
A.~Hermans, L.~Beyer, and B.~Leibe.
\newblock In defense of the triplet loss for person re-identification.
\newblock {\em CoRR}, abs/1703.07737, 2017.

\bibitem{MobileNetV1}
A.~G. Howard, M.~Zhu, B.~Chen, D.~Kalenichenko, W.~Wang, T.~Weyand,
  M.~Andreetto, and H.~Adam.
\newblock Mobilenets: Efficient convolutional neural networks for mobile vision
  applications.
\newblock {\em CoRR}, abs/1704.04861, 2017.

\bibitem{HuYYKCLH15}
G.~Hu, Y.~Yang, D.~Yi, J.~Kittler, W.~J. Christmas, S.~Z. Li, and T.~M.
  Hospedales.
\newblock When face recognition meets with deep learning: An evaluation of
  convolutional neural networks for face recognition.
\newblock In {\em {IEEE} International Conference on Computer Vision Workshop},
  pages 384--392, 2015.

\bibitem{LFWTech}
G.~B. Huang, M.~Ramesh, T.~Berg, and E.~Learned-Miller.
\newblock Labeled faces in the wild: A database for studying face recognition
  in unconstrained environments.
\newblock Technical Report 07-49, University of Massachusetts, Amherst, October
  2007.

\bibitem{CSM}
Q.~Huang, W.~Liu, and D.~Lin.
\newblock Person search in videos with one portrait through visual and temporal
  links.
\newblock In {\em Computer Vision - {ECCV} 2018 - 15th European Conference,
  Munich, Germany, September 8-14, 2018, Proceedings, Part {XIII}}, pages
  437--454, 2018.

\bibitem{Kenny05jointfactor}
P.~Kenny.
\newblock Joint factor analysis of speaker and session variability: Theory and
  algorithms.
\newblock Technical report, 2005.

\bibitem{odyssey/Kenny10}
P.~Kenny.
\newblock Bayesian speaker verification with heavy-tailed priors.
\newblock In {\em Odyssey 2010: The Speaker and Language Recognition Workshop},
  page~14, 2010.

\bibitem{KimKPR08}
M.~Kim, S.~Kumar, V.~Pavlovic, and H.~A. Rowley.
\newblock Face tracking and recognition with visual constraints in real-world
  videos.
\newblock In {\em {IEEE} Conference on Computer Vision and Pattern
  Recognition}, 2008.

\bibitem{DeepSpeaker}
C.~Li, X.~Ma, B.~Jiang, X.~Li, X.~Zhang, X.~Liu, Y.~Cao, A.~Kannan, and Z.~Zhu.
\newblock Deep speaker: an end-to-end neural speaker embedding system.
\newblock {\em CoRR}, abs/1705.02304, 2017.

\bibitem{CUHK03}
W.~Li, R.~Zhao, T.~Xiao, and X.~Wang.
\newblock Deepreid: Deep filter pairing neural network for person
  re-identification.
\newblock In {\em CVPR}, pages 152--159, 2014.

\bibitem{corr/LinZZWY17}
Y.~Lin, L.~Zheng, Z.~Zheng, Y.~Wu, and Y.~Yang.
\newblock Improving person re-identification by attribute and identity
  learning.
\newblock {\em CoRR}, abs/1703.07220, 2017.

\bibitem{corr/LiuJJQJYF17}
H.~Liu, Z.~Jie, J.~Karlekar, M.~Qi, J.~Jiang, S.~Yan, and J.~Feng.
\newblock Video-based person re-identification with accumulative motion
  context.
\newblock {\em CoRR}, abs/1701.00193, 2017.

\bibitem{cvpr/LiuWYLRS17}
W.~Liu, Y.~Wen, Z.~Yu, M.~Li, B.~Raj, and L.~Song.
\newblock Sphereface: Deep hypersphere embedding for face recognition.
\newblock In {\em {IEEE} Conference on Computer Vision and Pattern
  Recognition}, pages 6738--6746, 2017.

\bibitem{AlignedReID++}
H.~Luo, W.~Jiang, X.~Zhang, X.~Fan, Q.~Jingjing, and C.~Zhang.
\newblock Alignedreid++: Dynamically matching local information for person
  re-identification.
\newblock 2018.

\bibitem{IR}
C.~D. Manning, P.~Raghavan, and H.~Sch{\"{u}}tze.
\newblock {\em Introduction to information retrieval}.
\newblock Cambridge University Press, 2008.

\bibitem{conielecomp/MartinezPHS12}
J.~Mart{\'{\i}}nez, H.~P{\'{e}}rez{-}Meana, E.~E. Hern{\'{a}}ndez, and M.~M.
  Suzuki.
\newblock Speaker recognition using mel frequency cepstral coefficients
  {(MFCC)} and vector quantization {(VQ)} techniques.
\newblock In {\em International Conference on Electrical Communications and
  Computers}, pages 248--251, 2012.

\bibitem{interspeech/McLarenFCL16}
M.~McLaren, L.~Ferrer, D.~Cast{\'{a}}n, and A.~Lawson.
\newblock The speakers in the wild {(SITW)} speaker recognition database.
\newblock In {\em Interspeech}, pages 818--822, 2016.

\bibitem{MillerKS15}
D.~Miller, I.~Kemelmacher{-}Shlizerman, and S.~M. Seitz.
\newblock Megaface: {A} million faces for recognition at scale.
\newblock {\em CoRR}, abs/1505.02108, 2015.

\bibitem{VoxCeleb}
A.~Nagrani, J.~S. Chung, and A.~Zisserman.
\newblock Voxceleb: {A} large-scale speaker identification dataset.
\newblock In {\em Interspeech}, pages 2616--2620, 2017.

\bibitem{NagraniZ17}
A.~Nagrani and A.~Zisserman.
\newblock From benedict cumberbatch to sherlock holmes: Character
  identification in {TV} series without a script.
\newblock In {\em BMVC}, 2017.

\bibitem{SSH}
M.~Najibi, P.~Samangouei, R.~Chellappa, and L.~S. Davis.
\newblock {SSH:} single stage headless face detector.
\newblock In {\em {IEEE} International Conference on Computer Vision}, pages
  4885--4894, 2017.

\bibitem{icassp/PanayotovCPK15}
V.~Panayotov, G.~Chen, D.~Povey, and S.~Khudanpur.
\newblock Librispeech: An {ASR} corpus based on public domain audio books.
\newblock In {\em International Conference on Acoustics, Speech and Signal
  Processing}, pages 5206--5210, 2015.

\bibitem{RanjanCC17}
R.~Ranjan, C.~Castillo, and R.~Chellappa.
\newblock L2-constrained softmax loss for discriminative face verification.
\newblock {\em CoRR}, 2017.

\bibitem{SincNet}
M.~Ravanelli and Y.~Bengio.
\newblock Speaker recognition from raw waveform with sincnet.
\newblock {\em CoRR}, abs/1808.00158, 2018.

\bibitem{YOLO_V2}
J.~Redmon and A.~Farhadi.
\newblock {YOLO9000:} better, faster, stronger.
\newblock In {\em 2017 {IEEE} Conference on Computer Vision and Pattern
  Recognition, {CVPR} 2017, Honolulu, HI, USA, July 21-26, 2017}, pages
  6517--6525, 2017.

\bibitem{dsp/ReynoldsQD00}
D.~A. Reynolds, T.~F. Quatieri, and R.~B. Dunn.
\newblock Speaker verification using adapted gaussian mixture models.
\newblock {\em Digital Signal Processing}, 10(1-3):19--41, 2000.

\bibitem{eccv/RistaniSZCT16}
E.~Ristani, F.~Solera, R.~S. Zou, R.~Cucchiara, and C.~Tomasi.
\newblock Performance measures and a data set for multi-target, multi-camera
  tracking.
\newblock In {\em ECCV Workshops}, pages 17--35, 2016.

\bibitem{Garofolo93}
J.~S.~Garofolo, L.~Lamel, W.~M.~Fisher, J.~Fiscus, and D.~S.~Pallett.
\newblock Darpa timit acoustic-phonetic continous speech corpus cd-rom. nist
  speech disc 1-1.1.
\newblock 93:27403, 01 1993.

\bibitem{FaceNet}
F.~Schroff, D.~Kalenichenko, and J.~Philbin.
\newblock Facenet: A unified embedding for face recognition and clustering.
\newblock pages 815--823, 03 2015.

\bibitem{cvpr/SchroffKP15}
F.~Schroff, D.~Kalenichenko, and J.~Philbin.
\newblock Facenet: {A} unified embedding for face recognition and clustering.
\newblock In {\em {IEEE} Conference on Computer Vision and Pattern
  Recognition}, pages 815--823, 2015.

\bibitem{crowdhuman}
S.~Shao, Z.~Zhao, B.~Li, T.~Xiao, G.~Yu, X.~Zhang, and J.~Sun.
\newblock Crowdhuman: A benchmark for detecting human in a crowd.
\newblock {\em arXiv preprint arXiv:1805.00123}, 2018.

\bibitem{Shaver16}
C.~D. Shaver and J.~M. Acken.
\newblock A brief review of speaker recognition technology.
\newblock {\em Electrical and Computer Engineering Faculty Publications and
  Presentations}, 2016.

\bibitem{VGG}
K.~Simonyan and A.~Zisserman.
\newblock Very deep convolutional networks for large-scale image recognition.
\newblock {\em CoRR}, abs/1409.1556, 2014.

\bibitem{Sivic09}
J.~Sivic, M.~Everingham, and A.~Zisserman.
\newblock ``{W}ho are you?'' -- learning person specific classifiers from
  video.
\newblock In {\em CVPR}, 2009.

\bibitem{aaai/SongLLHC18}
G.~Song, B.~Leng, Y.~Liu, C.~Hetang, and S.~Cai.
\newblock Region-based quality estimation network for large-scale person
  re-identification.
\newblock In {\em AAAI}, 2018.

\bibitem{DeepID3}
Y.~Sun, D.~Liang, X.~Wang, and X.~Tang.
\newblock Deepid3: Face recognition with very deep neural networks.
\newblock {\em CoRR}, abs/1502.00873, 2015.

\bibitem{SunWT15}
Y.~Sun, X.~Wang, and X.~Tang.
\newblock Deeply learned face representations are sparse, selective, and
  robust.
\newblock In {\em {IEEE} Conference on Computer Vision and Pattern
  Recognition}, pages 2892--2900, 2015.

\bibitem{DeepFace}
Y.~Taigman, M.~Yang, M.~Ranzato, and L.~Wolf.
\newblock Deepface: Closing the gap to human-level performance in face
  verification.
\newblock In {\em {IEEE} Conference on Computer Vision and Pattern
  Recognition}, pages 1701--1708, 2014.

\bibitem{TaigmanYRW15}
Y.~Taigman, M.~Yang, M.~Ranzato, and L.~Wolf.
\newblock Web-scale training for face identification.
\newblock In {\em {IEEE} Conference on Computer Vision and Pattern
  Recognition}, pages 2746--2754, 2015.

\bibitem{Togneri2011AnOO}
R.~Togneri and D.~Pullella.
\newblock An overview of speaker identification: Accuracy and robustness
  issues.
\newblock {\em IEEE Circuits and Systems Magazine}, 11:23--61, 2011.

\bibitem{icassp/VarianiLMMG14}
E.~Variani, X.~Lei, E.~McDermott, I.~Lopez{-}Moreno, and
  J.~Gonzalez{-}Dominguez.
\newblock Deep neural networks for small footprint text-dependent speaker
  verification.
\newblock In {\em International Conference on Acoustics, Speech and Signal
  Processing}, pages 4052--4056, 2014.

\bibitem{eccv/VariorHW16}
R.~R. Varior, M.~Haloi, and G.~Wang.
\newblock Gated siamese convolutional neural network architecture for human
  re-identification.
\newblock In {\em European Conference on Computer Vision}, pages 791--808,
  2016.

\bibitem{eccv/VariorSLXW16}
R.~R. Varior, B.~Shuai, J.~Lu, D.~Xu, and G.~Wang.
\newblock A siamese long short-term memory architecture for human
  re-identification.
\newblock In {\em European Conference on Computer Vision}, pages 135--153,
  2016.

\bibitem{DBLP:conf/eccv/WangCLHCQL18}
F.~Wang, L.~Chen, C.~Li, S.~Huang, Y.~Chen, C.~Qian, and C.~C. Loy.
\newblock The devil of face recognition is in the noise.
\newblock In {\em European Conference on Computer Vision}, pages 780--795,
  2018.

\bibitem{Wang_MM18}
G.~Wang, Y.~Yuan, X.~Chen, J.~Li, and X.~Zhou.
\newblock Learning discriminative features with multiple granularity for person
  re-identification.
\newblock In {\em ACM Multimedia}, 2018.

\bibitem{iLIDS}
T.~Wang, S.~Gong, X.~Zhu, and S.~Wang.
\newblock Person re-identification by video ranking.
\newblock In {\em European Conference on Computer Vision}, pages 688--703,
  2014.

\bibitem{WolfHM11}
L.~Wolf, T.~Hassner, and I.~Maoz.
\newblock Face recognition in unconstrained videos with matched background
  similarity.
\newblock In {\em {IEEE} Conference on Computer Vision and Pattern
  Recognition}, pages 529--534, 2011.

\bibitem{yang2016wider}
S.~Yang, P.~Luo, C.~C. Loy, and X.~Tang.
\newblock Wider face: A face detection benchmark.
\newblock In {\em IEEE Conference on Computer Vision and Pattern Recognition
  (CVPR)}, 2016.

\bibitem{SelfAttention}
H.~Zhang, I.~J. Goodfellow, D.~N. Metaxas, and A.~Odena.
\newblock Self-attention generative adversarial networks.
\newblock {\em CoRR}, abs/1805.08318, 2018.

\bibitem{AlignedReID}
X.~Zhang, H.~Luo, X.~Fan, W.~Xiang, Y.~Sun, Q.~Xiao, W.~Jiang, C.~Zhang, and
  J.~Sun.
\newblock Alignedreid: Surpassing human-level performance in person
  re-identification.
\newblock {\em CoRR}, abs/1711.08184, 2017.

\bibitem{SpindleNet}
H.~Zhao, M.~Tian, S.~Sun, J.~Shao, J.~Yan, S.~Yi, X.~Wang, and X.~Tang.
\newblock Spindle net: Person re-identification with human body region guided
  feature decomposition and fusion.
\newblock In {\em {IEEE} Conference on Computer Vision and Pattern
  Recognition}, pages 907--915, 2017.

\bibitem{MARS}
L.~Zheng, Z.~Bie, Y.~Sun, J.~Wang, C.~Su, S.~Wang, and Q.~Tian.
\newblock {MARS:} {A} video benchmark for large-scale person re-identification.
\newblock In {\em European Conference on Computer Vision}, 2016.

\bibitem{Market1501}
L.~Zheng, L.~Shen, L.~Tian, S.~Wang, J.~Wang, and Q.~Tian.
\newblock Scalable person re-identification: {A} benchmark.
\newblock In {\em International Conference on Computer Vision}, pages
  1116--1124, 2015.

\end{thebibliography}
}
\end{document}